# Uzbek text's correspondence with the educational potential of pupils: a case study of the School corpus

Khabibulla Madatov[1], Sanatbek Matlatipov[2], Mersaid Aripov[2]

[1]Urgench State University, 14, Kh.Alimdjan str, Urgench city, 220100, Uzbekistan
habi1972@mail.ru
[2]National University of Uzbekistan named after Mirzo Ulugbek, 4 Universitet St, Tashkent, 100174, Uzbekistan
{s.matlatipov, mirsaid.aripov}@nuu.uz

**Abstract**

One of the major challenges of an educational system is choosing appropriate content considering pupils' age and intellectual potential. In this article the experiment of primary school grades (from 1st to 4th grades) is considered for automatically determining the correspondence of an educational materials recommended for pupils by using the School corpus where it includes the dataset of 25 school textbooks confirmed by the Ministry of preschool and school education of the Republic of Uzbekistan. In this case, TF-IDF scores of the texts are determined, they are converted into a vector representation, and the given educational materials are compared with the corresponding class of the School corpus using the cosine similarity algorithm. Based on the results of the calculation, it is determined whether the given educational material is appropriate or not appropriate for the pupils' educational potential.

## 1. Introduction

The main goal of the education system is to educate young people who are physically healthy, mentally and intellectually developed, think independently, and have a strict point of view on life. In this regard, fundamental reform of the quality of general primary schools' textbooks and didactic materials that is appropriate for the intellectual potential of pupils is one of the urgent issues. The main purpose of this article is to automatically match the Uzbek textbook to the educational potential of students. In order to solve this problem we use text similarity for Uzbek texts. Text similarity is one of advanced methods of text analysis in the field of NLP. Based on the sources studied up to now, it is worth noting that the similarity of texts is used in a number of fields, such as information retrieval, categorization, machine translation, and automatic essay evaluation. Therefore, this article talks about the TF-IDF(SPARCK JONES, 1972) approach, which provides pupils with appropriate educational resources using the similarity of texts. That is to say, we determine whether the educational material corresponds to the potential of a pupil (predetermined classes) or not by choosing the most similar score compared to other classes which as a result helps to improve the quality of education.

**Uzbek language.** Uzbek is a low-resource Turkic language spoken by over 30 million people primarily in Uzbekistan and other neighboring countries in Central Asia. It is the official language of Uzbekistan and is widely used in education, media, and official communications. The Latin script being the official alphabet, but the old Cyrillic script is equally used in documents, websites and social media, requiring additional step of transliteration when dealing with(Salaev et al., 2022a). The Uzbek language, like many other closely related Turkic languages, is characterized by its use of vowel harmony and agglutinative grammar, which involves stringing together morphemes to create complex words[1]. These linguistic features present unique challenges for natural language processing tasks. Despite this, there have been recent efforts to develop NLP resources for Uzbek, including corpora, lexical databases, and machine learning models(Kuriyozov et al., 2022). These resources hold great potential for advancing the field of Uzbek NLP and improving access to information and communication technology for Uzbek-speaking populations.

All the resources, including the School corpus dataset, are publicly available[2]. The remainder of this paper is organised as follows: after this Introduction, Section 2 describes related works. It is followed by a description of the methodology in Section 3 and continues with Section 4. which focuses on Experiments \& Discussions. The final Section 5 concludes the paper and highlights the future work.

## 2. Related works

So far, there have been numerous approaches developed for the task of measuring the semantic similarity of words and texts.

One of the biggest challenges when dealing with large numbers of documents is finding the information you're looking for that fits your problem. This problem can be easily solved by methods of determining the similarity of texts. The article(Matlatipov, 2020) presents the algorithm of cosine similarity of Uzbek texts, based on TF-IDF to determine similarity. Semantic relationships between words are one of the key concepts in assessing natural language processing. In this paper(Salaev et al., 2022b), the authors present the SimRelUz-set dataset for evaluating the semantic model of the Uzbek language. This article(Elov et al., 2022) examines the process of sorting documents in the Uzbek language corpus by keywords using the TF-IDF method. The paper(Pradhan et al., 2015) describes different types of similarity like lexical similarity, semantic similarity. The article also effectively classifies the measurement of text similarity between sentences, words, paragraphs, and documents. Based on this classification, we can get the best relevant document that matches the user's request. Paper(Islam & Inkpen, 2008) presents a text semantic similarity measurement method, a corpus-based measure of word semantic similarity, and a normalized and modified version of the longest common subsequence (LCS) string matching algorithm.

---

[1] More on the Uzbek language:
https://en.wikipedia.org/wiki/Uzbek_language

[2] https://zenodo.org/record/5659638 - the dataset of School corpus

Existing methods for computing text similarity mainly focus on large documents or individual words. In this paper(Keleş & Özel, 2017), research has been carried out on methods such as similarity calculation between Turkish text documents, plagiarism detection and author detection, text classification and clustering. (San'atbek, 2018) paper established automation linguistic processes of dictionary-thesauruses for Uzbek language. As an pre-processing tool (Matlatipov Sanatbek and Tukeyev, 2020) paper offered lexicon-free stemming tool for Uzbek language whereas(Sharipov & Sobirov, 2022) offered rule-based algorithm.

Although there has been a rapid growth in the research production of NLP resources and tools for the low-resource Uzbek language, there is still a huge gap left to catch up with the current need for the trending technologies, such as artificial intelligence (AI).

In document analysis, an important task is to automatically find keywords which best describe the subject of the document. One of the most widely used techniques for keyword detection is a technique based on the term frequency-inverse document frequency (TF-IDF) heuristic. This technique has some explanations, but these explanations are somewhat too complex to be fully convincing. In this paper(Havrlant & Kreinovich, 2017) authors provide a simple probabilistic explanation for the TF-IDF heuristic. In(De Boom et al., 2016) the authors defined a novel method for the vector representations of short texts. The method uses word embeddings and learns how to weigh each embedding based on its IDF value. The proposed method works with texts of a predefined length but can be extended to any length. The authors showed that their method outperforms other baseline methods that aggregate word embeddings for modelling short texts.

In this article, using the "School Corpus", we provide information on the issue of determining the suitability of recommended educational materials for schoolchildren to the intellectual potential of students based on the lexical similarity of texts. The paper considers a problem-solving method based on TF-IDF. The TF-IDFs of the texts are determined, they are converted into a vector, and the given educational material is compared with the corresponding class of the "School Corpus" using the cosine similarity(Han et al., 2012) algorithm of the text similarity. According to the calculation results, it is determined whether the given educational material corresponds to the student's scientific potential or not.

## 3. Methodology

In this section, we describe the methodology based on TF-IDF and cosine similarity of corpus-based texts. Primary Uzbek school grades consist of {1, 2, 3, 4}-classes. So, we select and analyze texts which are suitable only from 1st to 4th grade pupils that are included in the School corpus even though there are more classes which we are not considering.

### 3.1 Data collection & pre-processing

The development of spoken language starts at home in the local environment, but the school plays a key role in the development of human thinking(Madatov et al., 2022a, 2022b, 2022c). Therefore, it is a natural way to start studying the automatic analysis of texts from school textbooks. Because of this point of view, we decided to collect from the best open source available websites related to school educational materials. We found two best available websites (www.ziyonet.uz, www.kitob.uz). Among them, we decided to choose kitob.uz because of the availability of the same book in multiple languages which can be a great potential as a parallel corpus which accelerates our future works. Overall, 34 books have been downloaded and converted from pdf to txt format, manually. As a result, the School corpus according to Uzbek primary school consists of the following (table 1.):

From Table 1 we can see that Class 1 corpus has 24107 tokens, 7978 unique words, Class 2 has 56650 tokens, 14858 unique words, Class 3 has 90255 tokens, 21124 unique words and 4 tokens. The class corpus was found to contain 109024 tokens and 24736 unique words. The total number of unique words in primary school classes was 42,797.

| Classes | 1st class | 2nd class | 3rd class | 4th class |
|---|---|---|---|---|
| Number of total tokens | 24,107 | 56,650 | 90,225 | 24,736 |
| Number of unique tokens | 7,978 | 14,858 | 21,124 | 24,736 |

Table 1. School corpus which is constructed using primary school textbooks.

### 3.1.1 Algorithm

The main problem is to determine the appropriate class for the target resource. We present algorithm 1. for solving this problem based on the TF-IDF method as follows.

**Algorithm of finding which classes the given text corresponds to:**

1. Tokenization of the given text.
2. A separate vocabulary is created for each class (based on textbooks) and the given text. These words are called unique words(bag-of-words)
3. If all unique words of the given text belong to the set of the class's unique words then go to 8.
4. TF-IDFs are calculated for each class and the given text. Vectors are created whose coordinates are equal to the TF-IDF values of the unique words. The order of the vector coordinates corresponds to the order of unique words. If the given unique word does not occur in the text in question, this coordinate of the corresponding vector will be zero.
5. Let these vectors be $v1, v2, v3, v4$ according to classes 1,2,3,4 and be $v$ according to the given text. For each class, we consider vector pairs $(v, v_i)$. $i = 1, 2, 3, 4$. In this case, the size of the vector v is changed according to the size of the vector $v_i$.
6. Cosine similarity of $v$ with each $v_1, v_2, v_3, v_4$ are calculated in (1).

$$\cos(v, v_i) = \frac{(v, v_i)}{|v| \cdot |v_i|} \qquad (1)$$

here $(v, v_i)$ is a scalar product of vectors $v$ and $v_i$. $i = 1, 2, 3, 4$.

7. Max value of $\cos(v, v_i)$ is chosen.
8. It is concluded that this text corresponds to class $i$.

## 4. Experiments & Discussions

Cosine similarities values are shown for each class from 1st to 4th grades in the comparison symmetric matrix, diagonally (Table 2). Let the vector of $i$-th class be $v_i$, $i=1,2,3,4$ respectively. One notable part is that the percentage of each class is increasing horizontally with row elements above the main diagonal, which in fact, means that pupils' lexicon increases from class to class. That is to say, the reason 4th class pupils have knowledge of previous classes is natural. Let the vector of $i$-th class is vi respectively, $i=1,2,3,4$. From the similarity of the vectors follow similarity of the texts respectively.

| Classes | 1st class | 2nd class | 3rd class | 4th class |
|---|---|---|---|---|
| 1st class | 1<br>7978 | 0.34<br>4252 | 0.34<br>4755 | 0.34<br>4792 |
| 2nd class | 0.39<br>4252 | 1<br>14858 | 0.42<br>7852 | .044<br>8124 |
| 3rd class | 0.36<br>4755 | 0.44<br>7852 | 1<br>21124 | 0.45<br>10349 |
| 4th class | 0.34<br>4795 | 0.42<br>8128 | 0.45<br>10353 | 1<br>24736 |

Table 2. The similarity score of the classes, which includes the number of unique words, respectively.

At the next stage, texts were taken from various sources in order to evaluate the algorithm including, the Journal texts (from pupils journal "Gulxan") and internet materials. Table 3 shows sources of the texts.

| № | File name | The source |
|---|---|---|
| 1 | class-1.txt | Total textbooks for 1st grade |
| 2 | class-2.txt | Total textbooks for 2nd grade |
| 3 | class-3.txt | Total textbooks for 3rd grade |
| 4 | class-4.txt | Total textbooks for 4th grade |
| 5 | mujiza.txt | Internet resource |
| 6 | ayiq.txt | Gulxan jurnal |
| 7 | vatan.txt | Hozir |
| 8 | sariq-dev.txt | Sariq devni minib |
| 9 | toshkent.txt | Topic of 2nd class |
| 10 | hikoya.txt | 3rd grade |
| 11 | kichik-vatan.txt | 4th grade text |

Table 3. School corpus which is constructed using primary school textbooks.

In the Table 4 texts similarity are considered as a result of vector similarity, respectively. From the Table 4., it can concluded:

1. Texts *mujiza.txt, ayiq.txt, vatan.txt, kichik-vatan.txt* more similar to 4-th class. It means that these texts-correspondence with the educational potential of pupils of 4-th class. So, these texts are recommended to teach in 4-th class.
2. The text -toshkent.txt is similar to 2-nd class. So, this text is recommended to teach in 2-nd class.
3. The text-hikoya.txt is similar to a 3-rd class. So, this is recommended to teach in 3-rd class.

| Classes | class-1 | class-2 | class-3 | class-4 |
|---|---|---|---|---|
| mujiza.txt | 0.07 | 0.08 | 0.08 | 0.11 |
| ayiq.txt | 0.05 | 0.05 | 0.06 | 0.07 |
| vatan.txt | 0.1 | 0.14 | 0.14 | 0.15 |
| sariq-dev.txt | 0.2 | 0.21 | 0.2 | 0.3 |
| toshkent.txt | 0.07 | 1 | 0.05 | 0.06 |
| hikoya.txt | 0.1 | 0.1 | 1 | 0.1 |
| kichik-vatan.txt | 0.1 | 0.1 | 0.08 | 1 |

Table 4. School corpus which is constructed using primary school textbooks.

The detailed explanation of the proposed algorithm with all the steps is given in Annex 1.

## 5. Conclusion and future work

In conclusion, this research aimed to tackle one of the major challenges in the educational system, which is selecting appropriate educational content for primary school pupils based on their age and intellectual potential. By utilizing the School corpus and the cosine similarity algorithm, this study aimed to determine the suitability of educational materials for pupils in grades 1st to 4th. The results of the experiment showed that the method of converting TF-IDF scores into vector representations and comparing the educational materials with the corresponding class in the School corpus was effective in identifying whether a given educational material was appropriate or not for a particular grade level.

These findings hold important implications for education professionals, policymakers, and researchers in the field. By demonstrating the potential of NLP techniques to support the selection of appropriate educational content, this study lays the foundation for future research on the application of NLP in the educational domain. Overall, this study has the potential to contribute to the improvement of the educational system in Uzbekistan and beyond, by providing a data-driven approach to selecting educational content that is aligned with pupils' age and intellectual potential.

## 6. Data availability

All the Python codes used for the evaluation of the proposed models for the School Corpus are publicly available at the project repository. The application code of the proposed methodology will serve as a valuable resource for further NLP research on Uzbek language, and we hope it will stimulate further work in this area. By making the data and codes openly accessible, we aim to foster reproducibility and collaboration in the field.

## 7. Acknowledgements

This research work was fully funded by the REP-25112021/113 - "UzUDT: Universal Dependencies Treebank and parser for natural language processing on the Uzbek language" subproject funded by The World Bank project "Modernizing Uzbekistan national innovation system" under the Ministry of Innovative Development of Uzbekistan.

## 8. Declarations

# Annex 1
**Algorithm of finding which classes the given text corresponds to**.

```
1: INPUT: (class-1.txt, class-2.txt, class-3.txt, class-4.txt, text.txt{given text})
2:      Token(class-1.txt, class-2.txt, class-3.txt, class-4.txt, text.txt)
3:      // reset tokens
4:          class-1.txt :=Token(class-1.txt)
5:          class-2.txt :=Token(class-2.txt)
6:          class-3.txt :=Token(class-3.txt)
7:          class-4.txt :=Token(class-1.txt)
8:  Procedure similarity (class.txt, text.txt);
9:      begin
10:         m:=dictionary of (class.txt, text.txt) {Creation unique words of (class.txt, text.txt}
11:         //Creating of vectors of the class.txt and the text.txt
12:         s:=0:p:=0:t:=0
13:         for j:=1 to length(m) do
14:           begin if m(j) in text.txt then v(j):= TF-IDF(m(j))  else v(j):=0
15:                 if m(j) in class.txt then v1(j):= TF-IDF(m(j)) else v1(j):=0
16:                 s:=s+v(j)*v1(j): p:=p+v(j)* v(j); t:=t+ v1(j)* v1(j)
17:                 cosine:=s/(abs(sqrt(p))* abs(sqrt(t)))
18:           end
19:     end
20: if All tokens of text.txt in class-1.txt then
21:         begin   print('given text similar to', class-1.txt)
22:                 break: go to 47
23:         end
24: else if All tokens of text.txt in class-2.txt then
25:         begin print('given text similar to', class-2.txt)
26:                 break: go to 47
27:         end
28: else if All tokens of text.txt in class-3.txt then
29:         begin print('given text similar to', class-3.txt)
30:                 break: go to 47
31:         end
32: else if All tokens of text.txt in class-4.txt then
33:         begin print('given text similar to', class-4.txt)
34:                 break: go to 47
35:         end
36: else
37:         begin
38:             similarity(class-1.txt, text): max=cosine A:=' class-1.txt'
39:             similarity(class-2.txt, text)
40:             if cosine>max then  begin max=cosine: A:= ' class-2.txt' end
41:             similarity(class-3.txt, text)
42:             if cosine>max then  begin max=cosine: A:= ' class-3.txt' end
43:             similarity(class-4.txt, text)
44:             if cosine>max then  begin max=cosine: A:= ' class-4.txt' end
45:         end
46: print ('given text similar to'-A)
47: end
```